\documentclass{article}

\usepackage{microtype}
\usepackage{graphicx}
\usepackage{subfigure}
\usepackage{booktabs} 

\usepackage{hyperref}
\usepackage{amsmath}
\usepackage{amsfonts}
\usepackage{xcolor}
\usepackage{mdframed}
\usepackage{color,soul}
\usepackage{placeins}

\usepackage[accepted]{icml2018}

\icmltitlerunning{Practical Bayesian Optimization for Variable Cost Objectives}

\begin{document}

\twocolumn[
\icmltitle{Practical Bayesian Optimization for Variable Cost Objectives}




\begin{icmlauthorlist}
\icmlauthor{Mark McLeod}{to}
\icmlauthor{Michael A. Osborne}{to,goo}
\icmlauthor{Stephen J. Roberts}{to,goo}
\end{icmlauthorlist}

\icmlaffiliation{to}{Department of Engineering Science, University of Oxford}
\icmlaffiliation{goo}{Oxford-Man Institute of Quantitative Finance}
 
\icmlcorrespondingauthor{}{markm@robots.ox.ac.uk, mosb@robots.ox.ac.uk, sjrob@robots.ox.ac.uk}

\icmlkeywords{Machine Learning, ICML}

\vskip 0.3in
]



\printAffiliationsAndNotice{\icmlEqualContribution} 

\begin{abstract}
We propose a novel Bayesian Optimization approach for black-box objectives with a variable determining the tradeoff between evaluation cost and the fidelity of the evaluations. Our approach chooses this variable at each step under an entropy-based acquisition function. Further, we use a new approach to sampling support points, which we show to provide a better approximation to the desired distribution and allow faster construction of the acquisition function. This allows us to achieve optimization with lower overheads than previous approaches; our approach is also implemented for a more general class of problem. We show this approach to be effective on synthetic and real-world benchmark problems.
\end{abstract}
\section{Introduction}
Bayesian optimization aims to find the minimiser of a function, $x_{*} = \mathrm{argmin}_{x \in \mathbb{R}^{d}} f(x)$.
 This objective function is often multimodal and costly to evaluate, either due to the requirement for computation or an expensive physical experiment. We therefore wish to make as few evaluations as possible and are willing to expend non-trivial computation time on determining the next point to evaluate. There is a significant body of literature addressing this problem,  with applications including tuning the hyperparameters of many machine learning algorithms \cite{snoek2012practical, hernandez2014predictive,klein2015towards}, sensor set selection \cite{garnett2010bayesian} and  tuning the gait parameters of  bipedal \cite{Calandra2016}, quadrupedal \cite{lizotte2007automatic} and snake \cite{tesch2011using} robots.

A common approach to the problem of choosing the next step is to construct a model of the objective using all previous observations. In the above applications, the model used is a Gaussian Process (GP) \cite{rasmussen2006gaussian}, which provides a normally distributed estimate of the objective function at any point given training data. We then define an acquisition function, $\alpha(x)$, which quantifies how useful evaluation at $x$ is expected to be. Since this acquisition function can be evaluated using the GP estimate, rather than the expensive objective, we can perform a global search to find its maximiser and choose that point as our next evaluation of the objective. We provide a brief overview and propose an alternative metric for evaluating the performance of Bayesian Optimization methods in \S \ref{sec:bayesopt}.

In this work, we consider an extension of the expensive optimization problem by considering an additional variable, $s$, an environmental variable \cite{williams_sequential_2000}. The choice of this variable allows the objective to be evaluated with reduced accuracy at a lower cost. By carefully selecting a value for $s$ at each step, we wish to further reduce the cost of finding the minimum of the full cost objective. 
Previous approaches to the environmental variable setting \cite{klein2015towards,swersky2013multi} are based on the Entropy search method \cite{hennig2012entropy} and use a parametric kernel to take advantage of a strictly decreasing performance as the environmental variable allows an increased dataset size. We adapt Predictive Entropy Search \cite{hernandez2014predictive} to form a similar acquisition function over the environmental variable in \S \ref{sec:envvariables}, but use the Mat\'ern 5/2 kernel for more general application. We then show a novel method of selecting sample points to reduce overheads which is applicable to both Entropy Search and Predictive Entropy Search in \S \ref{sec:pmindraw}. Our more computationally efficient approach allows the practical optimization of objectives that have hitherto proved infeasible. We show results for common optimization benchmarks and real world applications in \S \ref{sec:expresults}.

\section{Bayesian Optimization}
\label{sec:bayesopt}
\subsection{Gaussian Processes}

Gaussian Processes are a  standard tool for performing inference for a function value with uncertainty given a set of observations \cite{rasmussen2006gaussian}. The model is characterized by a kernel function $k(x_{1},x_{2} \mid \theta)$ where $\theta$ are hyperparameters. Common choices are the squared exponential kernel $ A\exp(-0.5r^{2})$, where $r = \bigl({\sum_{d \in D} \frac{(x_{2d}-x_{1d})^{2}}{h_{d}}}\bigr)^{\frac{1}{2}}$, which models smooth, infinitely differentiable functions, and the Mat\'ern 3/2 and 5/2 kernels, $A(1+\sqrt{3}r)\exp(-\sqrt{3}r) $ and $A(1+\sqrt{5}r+\frac{5}{3}r^{2})\exp(-\sqrt{5}r) $, giving once and twice differentiable functions, respectively. In the experiments below, we have used the Mat\'ern 5/2 kernel which is a common choice in Bayesian optimization \cite{ klein2015towards, snoek2012practical, swersky2013multi}.

Ideally, we would marginalize over the hyperparameters $\lambda = [A, h]$ to obtain posterior estimates  
\begin{equation}
 p(x \mid D) = \int p(x \mid D, \lambda)p(\lambda) \mathrm{d}\lambda,
\end{equation}
given some prior $p(\lambda)$, to obtain the full posterior mean function. However, this is not usually an analytic function, so cannot be achieved exactly. Instead, we use slice sampling \cite{neal2003slice} of the hyperparameters to approximate the posterior as
\begin{equation}
 p(x \mid D) = \frac{1}{K}\sum_{k=0}^{K} p(x \mid D, \lambda_{k}) 
\end{equation}
 where the $K$ draws of the hyperparameter values have been made by slice sampling of their posterior likelihood given the data observed so far. This is a common choice for hyperparameter marginalization in Bayesian Optimization literature \cite{murray_slice_2010, snoek2012practical, swersky2013multi, swersky_freeze-thaw_2014}.

\subsection{Acquisition Functions}
A selection of acquisition functions are available. Common and simple choices include: Probability of Improvement over the current best observation \cite{lizotte2008practical,kushner1964new}; the Expectation of Improvement \cite{jones1998efficient,movckus1975bayesian}, and; a Lower Confidence Bound \cite{srinivas2009gaussian} on the GP. We consider the Entropy Search (ES)  acquisition function proposed by \citet{hennig2012entropy}, specifically the Predictive Entropy Search (PES) acquisition of \citet{hernandez2014predictive}, which is a fast approximation to ES.

In Entropy Search, the optimization is viewed not as finding locations of progressively lower values of the objective, but as gaining knowledge about the location of the global minimum. Specifically, prior belief about the location of the global minimum is represented as a probability distribution, $p(x_{*})$, the probability that $x_{*} = \mathrm{argmax}_{x}  f(x)$. We desire to maximize the relative entropy (KL-divergence) of this distribution from the uniform distribution. This occurs when $p(x_{*})$ is a delta located at $x_{\text{min}}$. Therefore the ES acquisition function selects points to produce greedy maximization of the mutual information between $x_{*}$ and $y$,
\begin{equation}
\begin{aligned}
 x_{n+1} &= \mathrm{argmax}_{x}\bigl( H\bigl[p(x_{*} \mid D_{n})\bigr] \\
& \qquad \qquad- \mathbb{E}_{x_{*}}\bigl[H[p(x_{*} \mid D_{n},x,y)]\bigr] \bigr).
\end{aligned}
\end{equation}
\subsection{Predictive Entropy Search}
The procedure for implementing ES requires considerable computation to achieve a good approximation to the ideal acquisition function. PES seeks a fast approximation to the ES acquisition. This is achieved by noting that the mutual information between the location $x_{*}$ and the next observed values $y_{n+1}$ is given by
\begin{equation}
\begin{aligned}
I&[x_{*},y_{n+1} \mid D_{n},x_{n+1}] \\
&= H[x_{*} \mid D_{n}] -\mathbb{E}_{y_{n+1}}\bigl[H[x_{*} \mid D_{n},x_{n+1},y_{n+1})]\bigr] \\
&= H[y \mid D_{n},x]-\mathbb{E}_{x_{*}}\bigl[H[y \mid D_{n},x,x_{*}]\bigr].
\end{aligned}
\end{equation}
That is, the information gained about the location of $x_{\text{min}}$ by evaluating at $x_{n+1}$ is equal to the information gained about the value of $f(x_{n+1})$ given the true location of the global minimum. The acquisition function, $\alpha(x)$ is the expected information gain about the value at $x_{n+1}$ given a true observation of the global minimum.
\begin{equation}
\begin{aligned}
\alpha(x_{n+1}) 
& = \Delta H \\
& = H[y_{n+1} \mid x_{n+1}, D_{n}] - H[y_{n+1} \mid D_{n},x_{n+1},x_{*}]\\
\end{aligned}
\end{equation}
 To implement this variation a draw $x_{d}$ is from the distribution over the location of minimum given the current model, and at this location the minimizing conditions,
\begin{align}
f(x_{d}) & \leq \min(Y_{n}),  \\
\frac{\partial f(x_{d})}{\partial x_{i}} &= 0 \quad \forall i \qquad \mathrm{and}\\
\frac{\partial^{2} f(x_{d})}{\partial x_{i} \partial x_{j}} &  
\begin{cases} = 0  &i \neq j\\
\geq 0 &i = j
\end{cases}
\end{align}
 are imposed. Expectation Propagation \cite{minka2001expectation} is used to achieve a Gaussian approximation to the inequalities. The change in entropy of $y$ at a candidate $x_{n+1}$, averaged over draws of the minimum, is a good approximation to the ideal objective. We use a bespoke GP package\footnote{https://github.com/markm541374/gpbo} to accommodate observation and inference of first and second derivatives.

\subsection{Evaluation of Performance}
\label{sec:evaluation}
\begin{figure}
\centering
\includegraphics[width= \columnwidth]{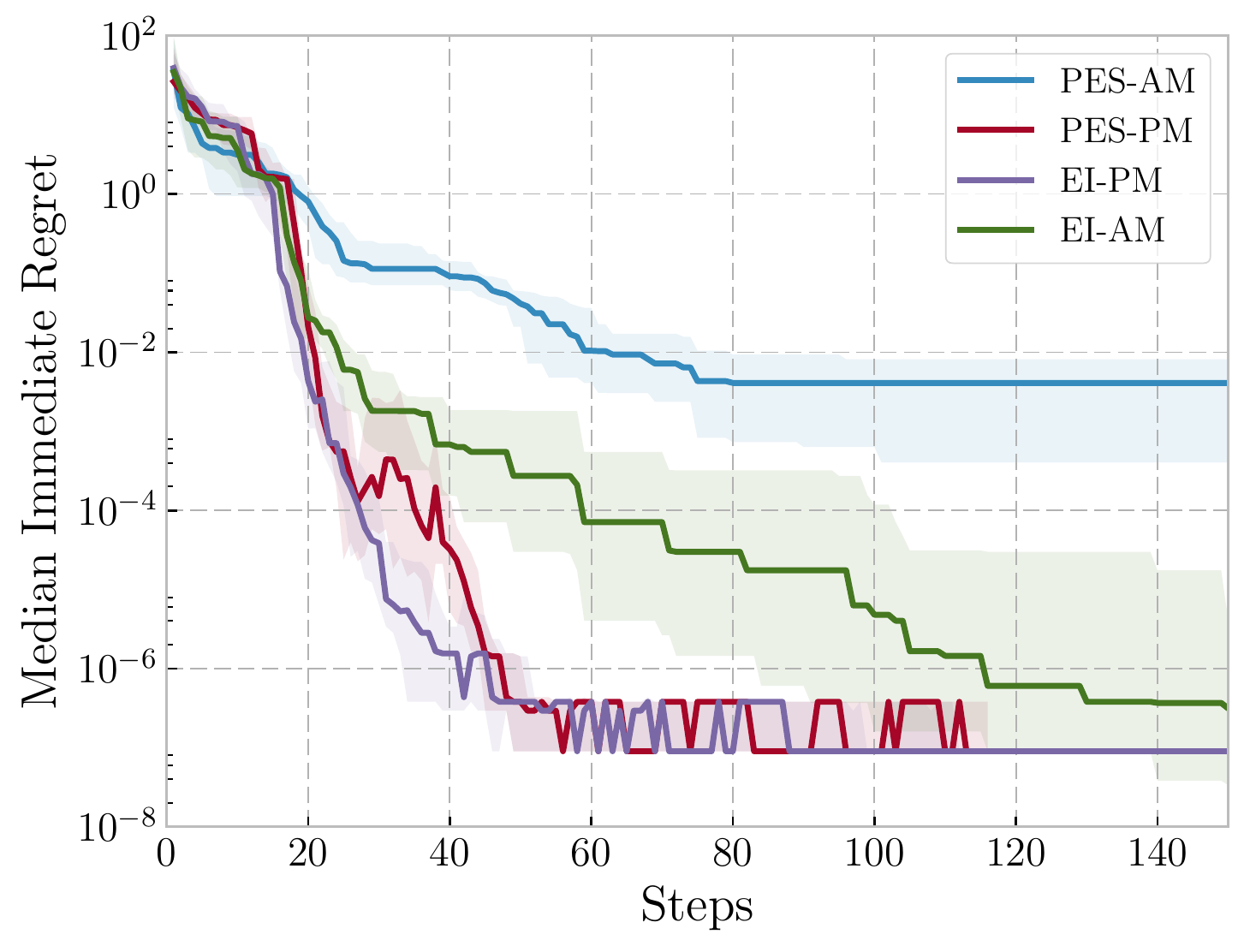}

\caption{ Performance on the Branin function of EI and PES under reporting of the minimum observed value (ArgMin) and the posterior minimum (PM) value. The median (solid line) and interquartile range (shaded) of 25 runs are shown. Performance is similar initially but reporting the posterior minimum rather than the minimum observed value leads to substantial lower errors, particularly with PES.}
\label{argpost}
\end{figure}

Since the point determined by the acquisition function is different from the posterior minimum the sequence of points evaluated does not represent the best guess for the global minimizer at each step. When using acquisition functions based on the values observed, such as expected improvement, these points do still provide good results. However, when using entropy-based methods, we find the points evaluated tend to be far from the posterior minimum. We therefore propose a greedy evaluation at the posterior minimum as the final step of optimization. In the experiments below, we perform this evaluation offline at each step and report the immediate regret (IR) that would have been returned if that step had been the last, before continuing according the regular optimization policy. The difference in performance by taking this approach is illustrated in Figure \ref{argpost}. We argue that all information theoretic means of Bayesian Optimization should be evaluated according to this metric.

\section{Environmental Variables}
\label{sec:envvariables}
We now consider an extension of the classic Bayesian Optimization setting. At each evaluation of the objective function we specify an additional parameter, named by \citet{klein2015towards} as the environmental variable. This parameter allows a tradeoff between accurate, but more expensive, and cheap, but poor, evaluations. We define the environmental variable as $0\leq s \leq 1$ where $s=0$ yields the true objective and cost is expected to decrease towards $s=1$. We desire an acquisition function that will provide the most cost efficient information gain at each step.

\subsection{Augmented space}
If variation of the environmental variable causes change in the mean of the response then, as in \citet{klein2015towards}, we model the black box as a $d+1$ dimensional function in $\mathbb{R}^{d} \times S,\ s \in S = [0,1] \subset \mathbb{R} $, such that $s=0$ corresponds to the true objective. This is the case in many scenarios. For example, in optimizing the parameters of a classifier, the performance is expected to be worse with fewer training data, or, in fitting the hyperparameters of a Gaussian process, the likelihood on a subset of the data is not expected to have the same value as on the full dataset. We then wish to define an acquisition function over the full space such that we learn about the minimum in the $s=0$ plane (that which corresponds to the true, uncorrupted, objective). Rather than the Entropy Search method as in \cite{klein2015towards}, we make use of Predictive Entropy Search, improved with a novel, faster, method of evaluating the expected change in entropy. We model the response of the objective as a GP over the augmented $d+1$ dimensional space, with a Mat\'ern 5/2 kernel, but rather than using a spectral decomposition to draw from $p(x_{*})$ we use sampling in the $s=0$ plane to draw support points, and draw from the posterior on these points. The method is detailed in Section \ref{sec:pmindraw}.

To optimize a wide range of objectives we are not able to guarantee that the minimum of the objective (for $s=0$) is the global minimum in the augmented space spanning all $s$. For this reason, we do not include the global minimum constraint used in the original specification of PES \cite{hernandez2014predictive} since the values reported at reduced cost may be lower as well as higher than the full cost result. For the same reason we use the $D+1$ dimensional Mat\'ern 5/2 product kernel to model the objective without any further assumptions on behaviour due to the environmental variable.

\subsection{Overhead adjustment}

In this section, we present a novel acquisition function over environmental variables. In the usual setting for Bayesian Optimization, the overhead computational cost of optimizing the acquisition function is considered negligible compared to evaluations of the true objective. When the option of a larger number of less expensive evaluations is made available, this may no longer be the case, particularly considering the poor scaling of Gaussian Processes (typically $\mathcal{O}(n^{3})$) as additional points are added. \citet{klein2015towards} use an acquisition function of the form
\begin{equation}
\alpha = \frac{\Delta H}{c(x,s)+c_{\text{over}}},
\end{equation}
where $c_{\text{over}}$ is the time for the previous step to choose the next point to evaluate and $c(x,s)$ is the GP posterior mean of the log evaluation cost conditioned on the evaluations observed so far using MAP hyperparameters of the Mat\'ern 5/2 kernel. Since the overhead grows substantially over the course of optimization, we prefer to use an estimate of the average overhead between the current and final steps.

We model the overhead as growing according to a power plus constant rule,
\begin{equation}
\hat{c}_{\text{over}}(n \mid \theta) = \theta_{0} + \theta_{1}n^{\theta_{2}} + \epsilon, \quad \epsilon \sim \mathcal{N}(0,\theta_{3}^2),
\end{equation}
where $n$ is the index of the step, and with independent Gamma priors on $\theta_{i}$. Given a remaining optimization budget $B$, our modified acquisition function is
\begin{equation}
\alpha = \frac{\Delta H}{c(x,s)+\frac{1}{N}\sum_{n=0}^N \hat{c}_{\text{over}}(n \mid \hat{\theta})}
\end{equation}
where $N$ is the greatest value such that
\begin{equation}
B \geq \sum_{n=0}^N \hat{c}_{\text{over}}(n \mid \hat{\theta})+c_{\text{evaluation}} \quad.
\end{equation}
Here $\hat{\theta}$ is the maximum-a-posteriori estimate given the overheads observed so far and the current step is considered to be step zero.

This change causes the algorithm to prefer slightly more expensive evaluations than otherwise, particularly when a large number of evaluations remain, which we find to improve performance. 

\section{Fast Draws from the Posterior of the Minimizer}
\label{sec:pmindraw}
We now present a novel sampling strategy for PES that renders our method computationally efficient.
The original formulation of PES in \citet{hernandez2014predictive} makes use of Bochner's theorem to obtain approximate draws from the Gaussian process which can easily be minimized to obtain a draw from the posterior distribution of the global minimizer. We prefer the alternate method of generating draws, proposed in \cite{hennig2012entropy}, of drawing support points from some distribution $q(x)$, which is similar to $p(x_{*})$, then making draws from the GP posterior to provide samples of $p(x_{*})$. This process does not place any requirements on the kernel used, unlike the original proposal which requires a stationary kernel (or an approximation in the case of a non-stationary kernel). As noted by \citet{hennig2012entropy}, any $q$ with non-zero support over the search domain may be used, with more samples of $q$ being required to obtain good results if it is not similar to $p$.

Slice sampling over the EI of the GP posterior is the suggested method of drawing support points in \cite{hennig2012entropy}. However, the evaluation of EI requires $O(n^{2})$ inference and is performed many times for each point produced by slice sampling. This is further increased by the practice of discarding a burn-in period and subsampling the resulting sequence. This represented a large portion of the runtime in our implementation. As any $q$ can be used, we seek an alternative from which we can draw points with lower computational overhead using a Weighted mixture of Local Hessian matrices (WLH).

To achieve a fast approximation for $p$ we note that the probability of a point being stationary (either a local extremum or an inflection point) is equal to the probability 
\begin{equation}
p_{\text{stat}} = \prod_{i=0}^{D} P\left( \frac{\partial f}{\partial x_{i}} = 0 \right)
\end{equation}
 by definition. Further, the local minima, $x^{l}_{i}$, of the posterior mean are local maxima of $p_{\text{stat}}$. We can find these points easily by performing local searches on the posterior mean from random start points, following which no further iteration with the GP model is required  regardless of the number of support points desired.
Given a candidate local minima $x^{c}_{i}$ we infer the GP mean and covariance of all elements of the gradient, $g_{\mu}, \Sigma_{g}$, and Hessian, $H_{\mu}, \Sigma_{H}$, at that point. We then make a second order Taylor approximation centred at the candidate minimum,
\begin{equation}
f = \frac{1}{2} z^{T}Hz + z^{T}g+c,
\end{equation}
where $z=x-x^{c}_{i}$. We make the further assumption that $\mu_{H} \gg \Sigma_{H}$ for all elements of $H$, therefore the Hessian is treated as constant $H=\mu_{H}$. The local minimum under this model is located at
\begin{equation}
\begin{aligned}
\frac{\partial f}{\partial z} &= \underline{0}\\
z &= H^{-1}g.
\end{aligned}
\end{equation}
Since $g$ is Normally distributed, and zero at $x^{c}_{i}$ by definition, the posterior distribution for the local minimum under the Taylor approximation is
\begin{equation}
p(x^{l}) = \mathcal{N}\bigl( x^{c}, H^{-1} \Sigma_{g} H^{-T}  \bigr).
\end{equation}


\begin{figure*}
\centering

\includegraphics[width=0.9\columnwidth]{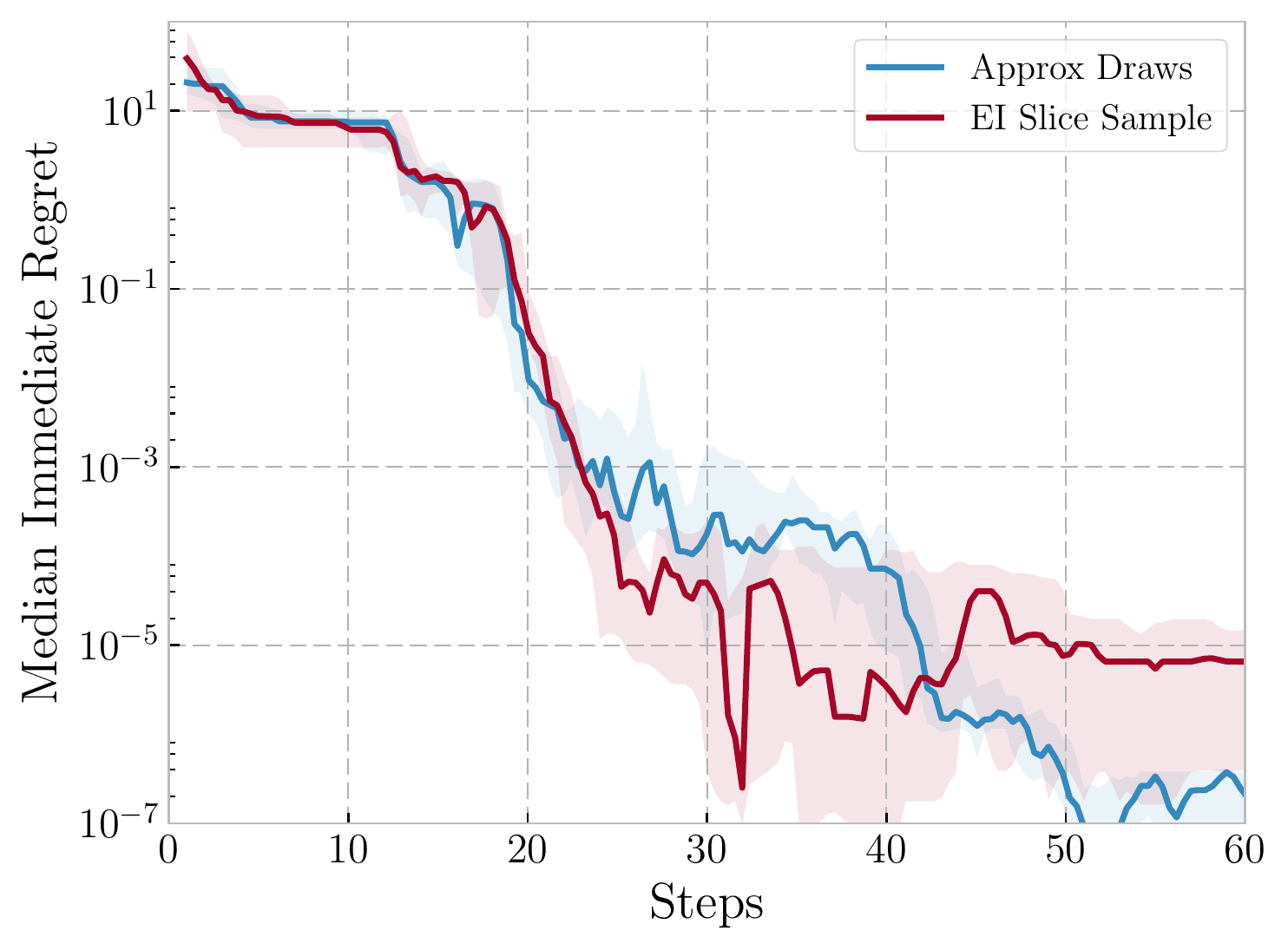}
\includegraphics[width=0.9\columnwidth]{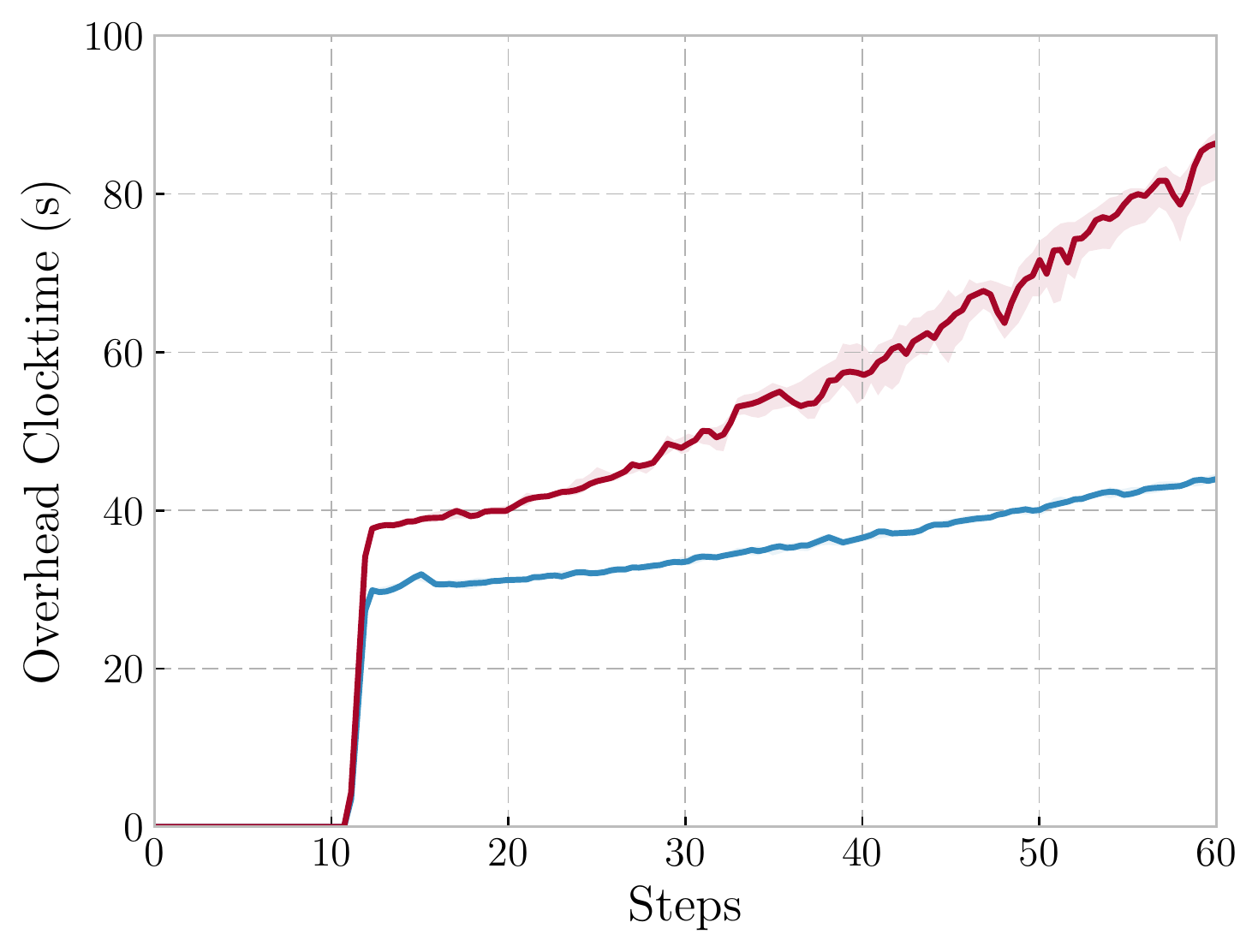}
\caption{Overhead of optimizing the acquisition function and immediate regret for an offset Branin objective with quadratic cost. Slice sampling under EI (red) and our sum of Gaussians approximation (blue) are shown. Performance remains similar while overhead costs are significantly reduced.  }
\label{overhead}
\end{figure*}

We now combine draws from the distributions of each separate local minimum according to a weight vector $w$. The ideal $w$ would be the respective probabilities of each local minimum being the global minimum.  Unfortunately even the correct weights, only considering the $z=0$ points, are intractable. To obtain a rough approximation, we consider the mean and variance $c_i, \sigma_i^2$ at each $x_i^l$  and let 
\begin{equation}
\begin{aligned}
\bar{w} &=p\left( \mathcal{N}(c_i,\sigma^2_i)<\mathcal{N}(c_*,\sigma^2_*) \right) \\
w &= \frac{\bar{w}}{\sum_i  \bar{w_i}} \,,
\end{aligned}
\end{equation}
where $c_*, \sigma_*^2$ are the minimum $c_i$ and corresponding variance. This has the desirable properties that the local minimum with the lowest expected value has the greatest weight, and for all others the weight increases monotonically with the expected value, and decreases monotonically with the variance.

We therefore can express $q(x)$, the distribution we use to draw support points, as 
\begin{equation}
q = Z \sum_{i}^{m} w_i \mathcal{N}_{\text{bounded}}( x^{c}_{i}, H_{i}^{-1} \Sigma_{gi} H_{i}^{-T}) ,
\end{equation}
where there are $m$ local minima and $Z$ is a normalizing constant.  The use of $\mathcal{N}_{\text{bounded}}$ denotes that the distribution is clipped within the search domain. We can take draws from this distribution trivially, since it is only a weighted combination of Normal distributions, and so can generate as many samples as desired with no additional interaction with the expensive GP model.

\subsection{Validation}
Since we do not have access to the true distribution $p(x_*)$ we compare our new method of generating support points by evaluating a set of quality metrics offline on a selection of optimizations using the PES acquisition function. For each step of optimization we generate $1000$ support points using each method and then take $10,000$ samples from the GP posterior on these points. Our method is compared to slice sampling over EI and LCB. Sampling from the uniform distribution is included as a baseline. We then consider four quality metrics for these samples.
\begin{description}
\item[KL-divergence] The distribution of the posterior minimum when confined to $m$ support points is an $m$-dimensional multinomial distribution, which if the support points have been drawn from $p(x_*)$ will have uniform probability for each component. Using a Dirichlet prior we can find the parameters which maximize the posterior likelihood conditioned on the observed draws. The KL-divergence from this uniform to the multinomial with these parameters provides a measure of how similar our approximating distribution is to $p(x_*)$.


\item[Unused points] If we take $N=10,000$ samples from a discrete uniform distribution with $m=1000$ values the expected number of values not drawn is less than one. If a method has a significant number of unused points this indicates a large number of our support points have a very low value for $p(x_*)$, so cost us additional computation time with no benefit.

\item[Time] Reducing execution time, $t$ is, of course, our primary objective.

\item[Rate of useful point production] This is the rate at which points making a useful contribution to our support set are generated. We define a useful point as any that is selected as the global minimum at least $\frac{N}{10m}$ times from the $N$ samples of the GP posterior over the size $m$ support set. That is, at least one tenth as often as if all support points were truly drawn from $p(x_*)$. The rate of production of useful points is therefore $r = \frac{n-1}{t}$, where $n$ is the number of support points that have fulfilled this criterion (we use $n-1$ in the numerator since a support set containing single point which is always selected does not convey useful information).
\end{description}

The results of these comparisons are shown in Table \ref{ds_table}. As expected uniform draws perform poorly, with high KL divergence and almost all points unused. Although this method is technically superior in execution time and the rate of generation of useful support points the requirement to take an order of magnitude more support points to obtain a similarly sized useful set would be prohibitively expensive in future steps, more than cancelling out the gains in obtaining them. Our method outperforms the slice sampling methods in almost all tests, with similar performance over all four methods in the remaining two.

\begin{table}
\begin{center}
\begin{tabular}{c c c c c  } 
 \toprule

Objective & Uniform& EI &LCB& WLH\\
 &\multicolumn{4}{c}{KL-divergence} \\ \cmidrule(r){2-5}

Branin 2d   &4.53 &3.44 &3.67 &\textbf{0.425}  \\
Hartmann 3d &4.92 &2.87 &4.82 &\textbf{0.451}  \\
Draw 4d     &1.75 &0.854 &1.8 &\textbf{0.738}  \\
Hartmann 6d &1.83 &0.715 &2.13 &\textbf{0.627}  \\

%
 \addlinespace
  &\multicolumn{4}{c}{Percent Useful points} \\ \cmidrule(r){2-5}

Branin 2d   &17.3 &31.8 &24.7 &\textbf{77.1} \\
Hartmann 3d &13.9 &36.1 &17.1 &\textbf{73.3}  \\
Draw 4d     &69.0 &\textbf{89.5} &72.8 &68.6  \\
Hartmann 6d &66.7 &\textbf{88.2} &65.8 &77.9  \\

 \addlinespace
   &\multicolumn{4}{c}{Time} \\ \cmidrule(r){2-5}

Branin 2d   &0.000184 &20.8 &37.1 &\textbf{0.489}  \\
Hartmann 3d &0.000216 &39.1 &104 &\textbf{1.27}  \\
Draw 4d     &0.000248 &68.2 &147 &\textbf{2.93}  \\
Hartmann 6d &0.000309 &146 &230 &\textbf{8.98}  \\
\addlinespace
&\multicolumn{4}{c}{Rate of Useful Point Generation} \\ \cmidrule(r){2-5}

Branin 2d   &1.24e+06 &34 &12.1 &\textbf{1.69e+03}  \\
Hartmann 3d &5.37e+05 &12.3 &1.73 &\textbf{745}  \\
Draw 4d     &2.2e+06 &15.7 &4.44 &\textbf{255} \\
Hartmann 6d &2.26e+06 &7.78 &3.31 &\textbf{131}  \\
\addlinespace
 \bottomrule
\end{tabular}
\end{center}
\caption{Performance of methods of drawing support. Best performance shown in bold on each row, excluding uniform support which is included only as a reference. Results are averaged over 16 optimizations up to 50 iterations on each objective. ``Draw 4d'' denotes objectives drawn at random from the 4 dimensional Mat\'ern 5/2 kernel.Our weighted local Hessians approximation achieves the best performance on 14 of the 16 tests.}
\label{ds_table}
\end{table}

\section{Experiments}

\label{sec:expresults}
We compare our method (EnvPES) to Expected Improvement (EI), Predictive Entropy Search (PES) and FABOLAS. Our GP package is used for PES and EI, the implementation of FABOLAS is provided in the RoBO package\footnote{https://github.com/automl/RoBO/}. We modify these implementations to return the posterior mean minimizer as in \S\ref{sec:evaluation}, and further modify FABOLAS to use the Mat\'ern 5/2 kernel over the environmental variable rather than the parametric form used in the original since the monotonic assumption does not necessarily hold in our experiments. We show that we are able to match or exceed the performance of existing methods over a selection of objectives.

For initialization we follow \citet{klein2015towards} by evaluating a set of values of the environmental variable for each random draw from $x$ rather than each evaluation being independent. We choose $s=0.5,0.75,0.875$ and use twenty total evaluations in the initialization dataset.

\subsection{In-model test}
\begin{figure*}[h]
\centering
\includegraphics[width=\columnwidth]{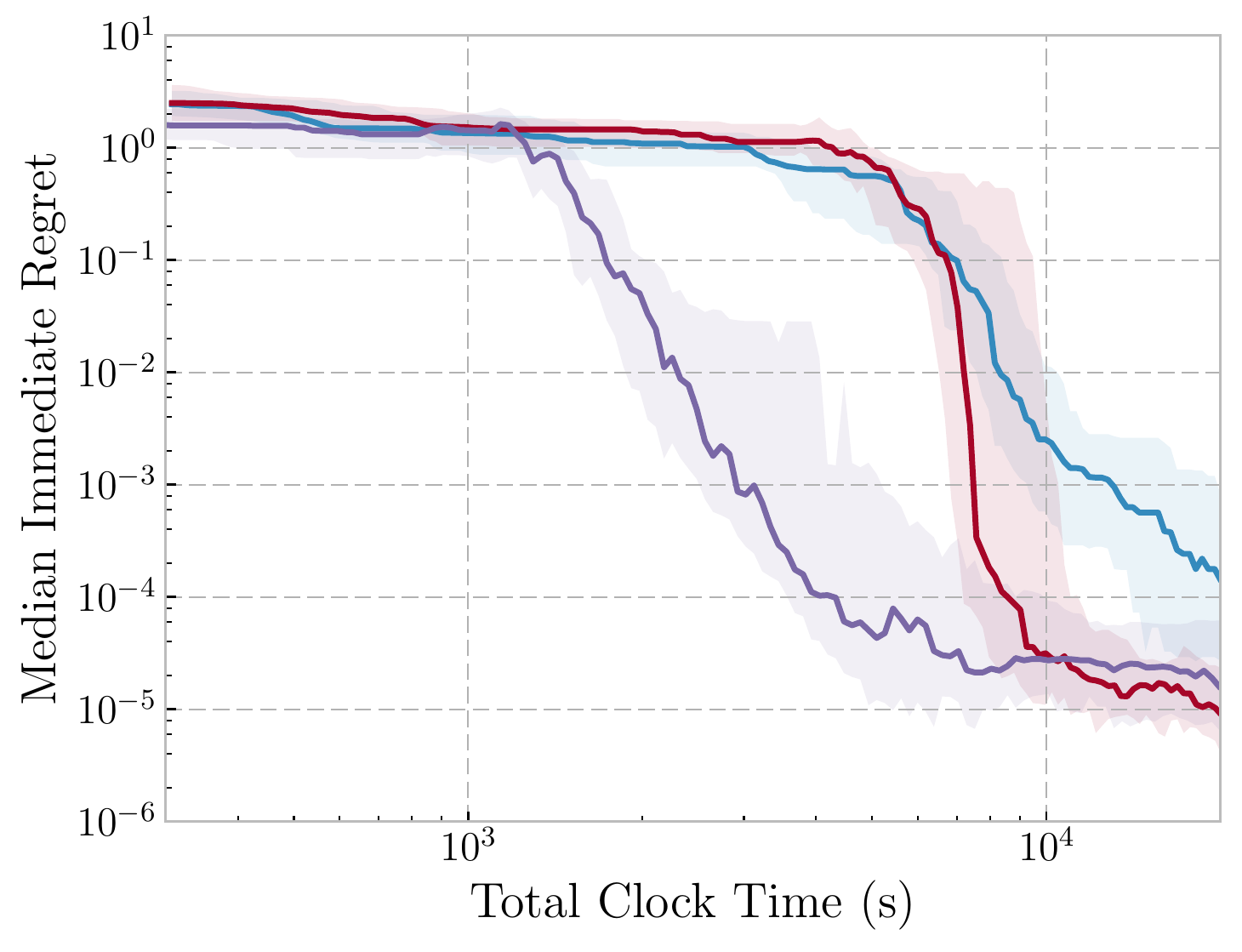}
\includegraphics[width=\columnwidth]{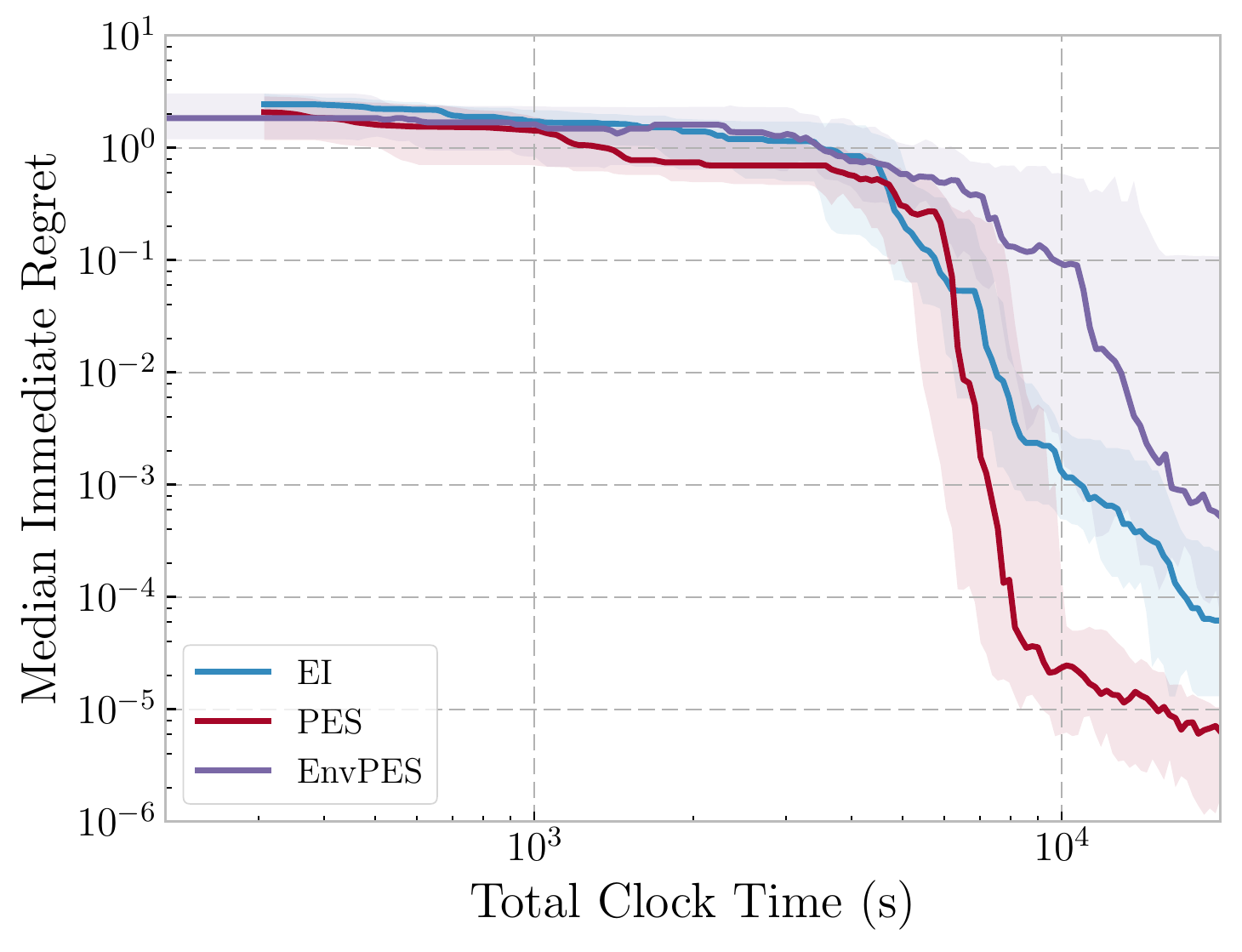}

\caption{ Performance of EnvPES (green), PES (red) and Expected Improvement (blue) on draws from the Mat\'ern kernel. The median (solid line) and interquartile range (shaded) of optimizing 40 draws are shown. Left: The objective and cost function had $l_{c}=3$ and $l_{ev}=1.5$ which allows very good performance as the cost decays quickly while the approximate objective remains very similar to the true objective. Substantial improvement is therefore available using less expensive evaluations. Right: The objective and cost function had $l_{c}=1$ and $l_{ev}=0.4$ which induces poor performance as the cost of evaluation does not reduce much and evaluations at increasing $s$ are only loosely linked to the true objective. Evaluations cannot be made as cheaply compared to the full cost as the previous example and convey little information about the true objective. EnvPES is no longer able to perform as well as other methods since it needs to learn a more complex model.}
\label{draws}
\end{figure*}

To illustrate expected performance we use draws from the Mat\'ern 5/2 kernel as objective function. The characteristic lengthscale is set to $0.3$ and the search domain is $[-1,1]^{2}$. We simulate the environmental variable as an additional dimension of the objective over $[0,1]$ with characteristic length $l_{ev}$ and the cost as an exponential $\exp(- l_{c} s)$. We show results for advantageous and adversarial values of $l_{c}$ and $l_{ev}$ in Figure  \ref{draws}. This shows the potential gains of making use of the environmental variable in a suitable scenario, while retaining reasonable performance otherwise.

\subsection{Off-model test}
\subsubsection{Common Synthetic Functions}
\begin{figure*}
\centering

\includegraphics[width=0.32\textwidth]{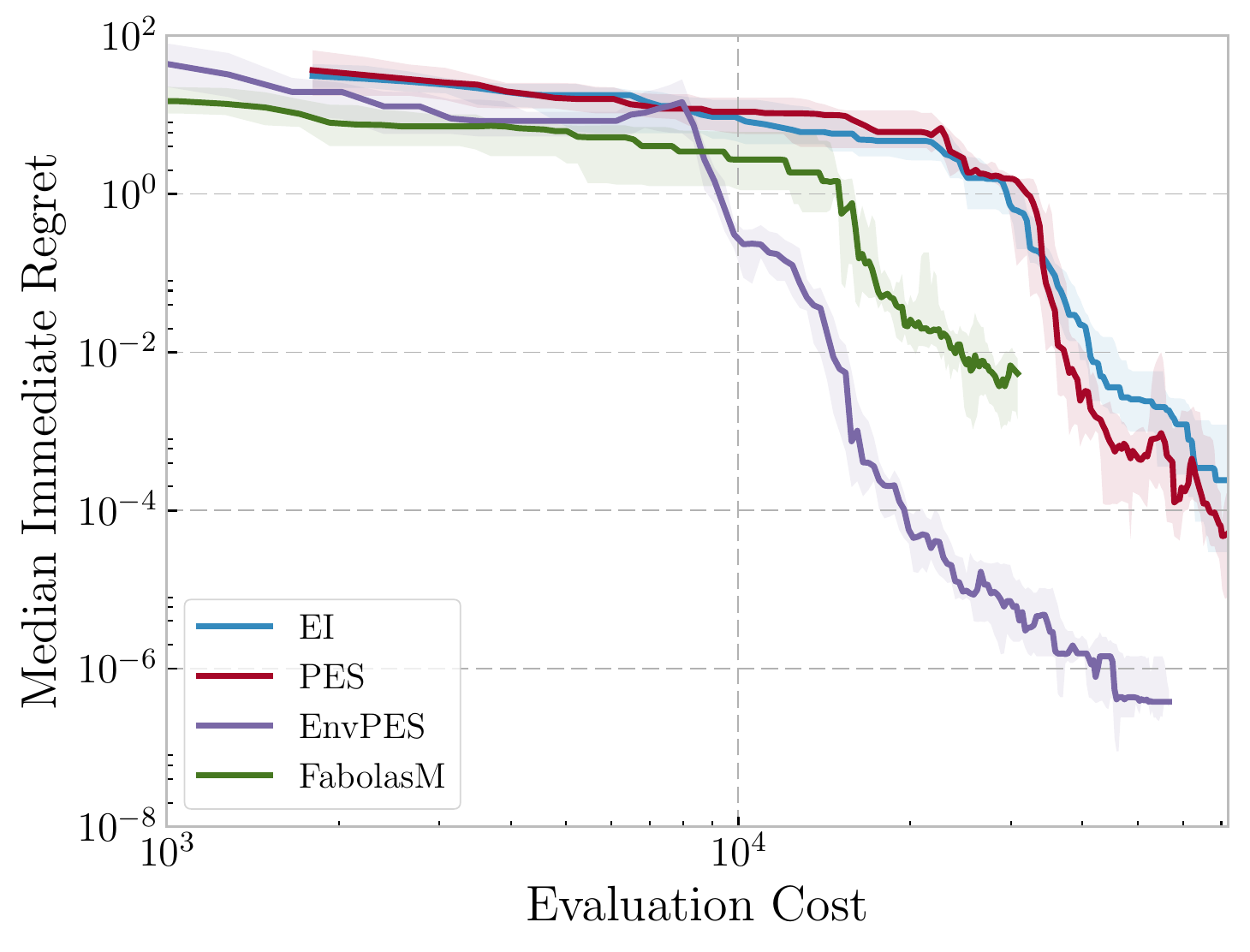}
\includegraphics[width=0.32\textwidth]{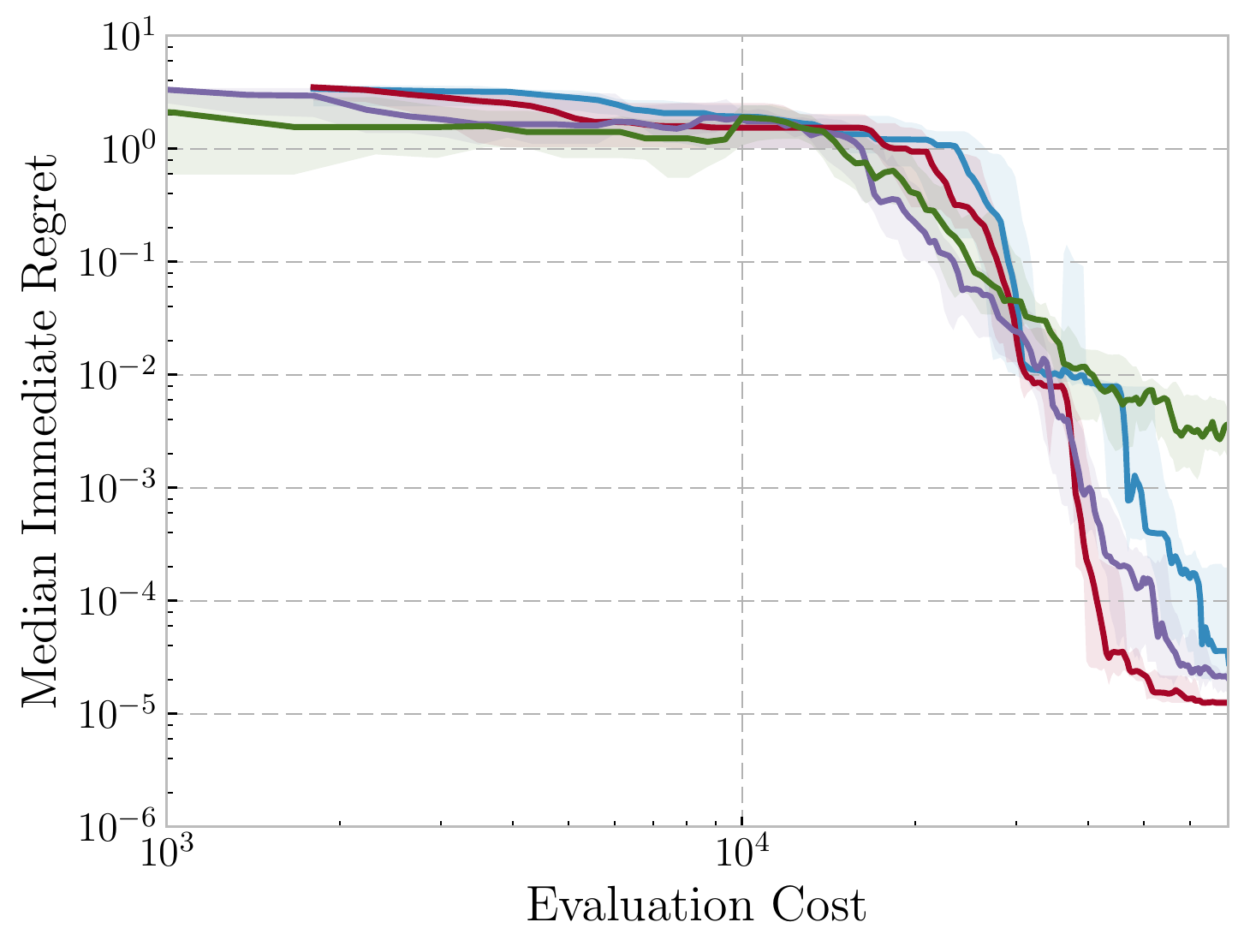}
\includegraphics[width=0.32\textwidth]{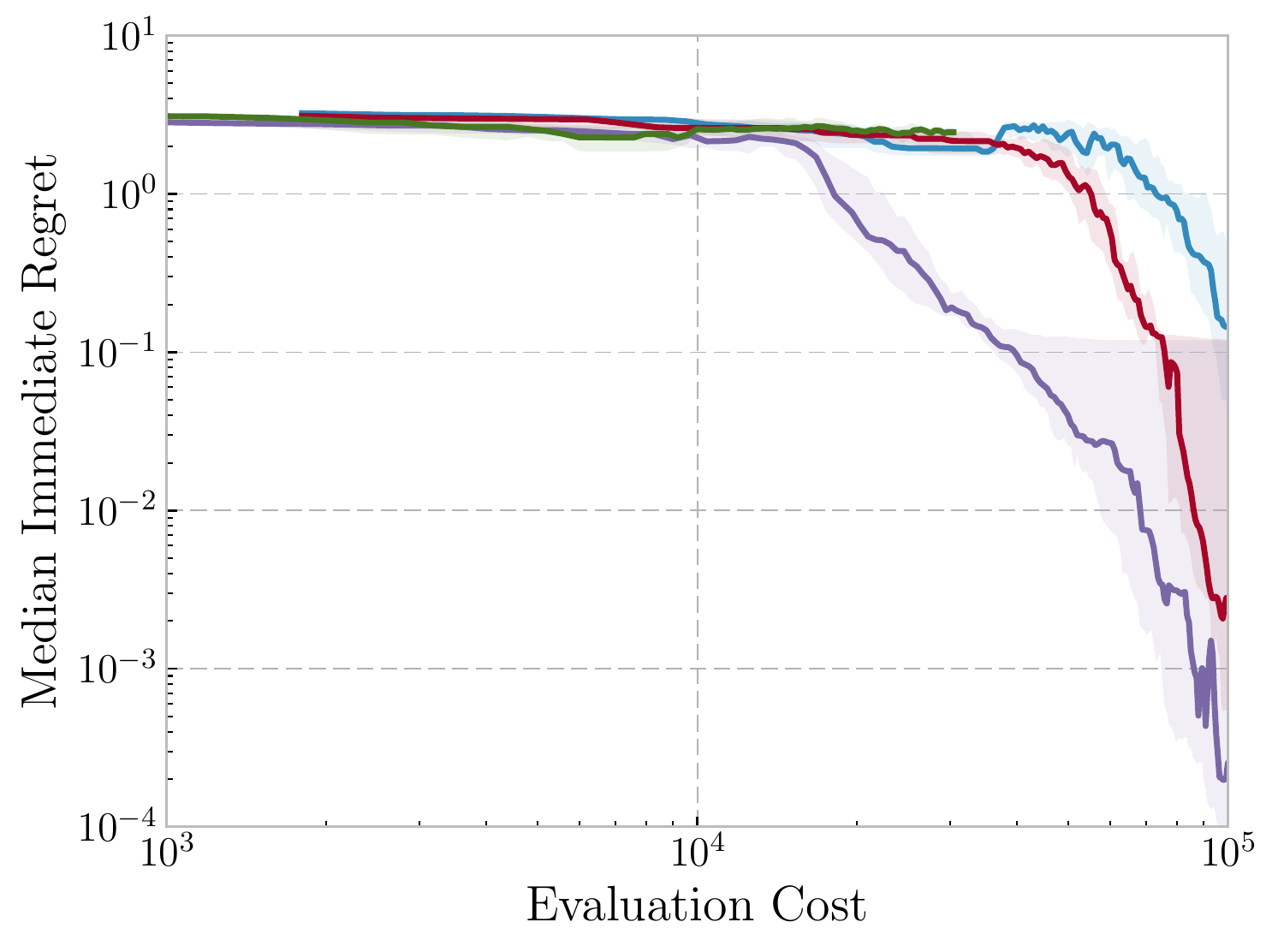}

\includegraphics[width=0.32\textwidth]{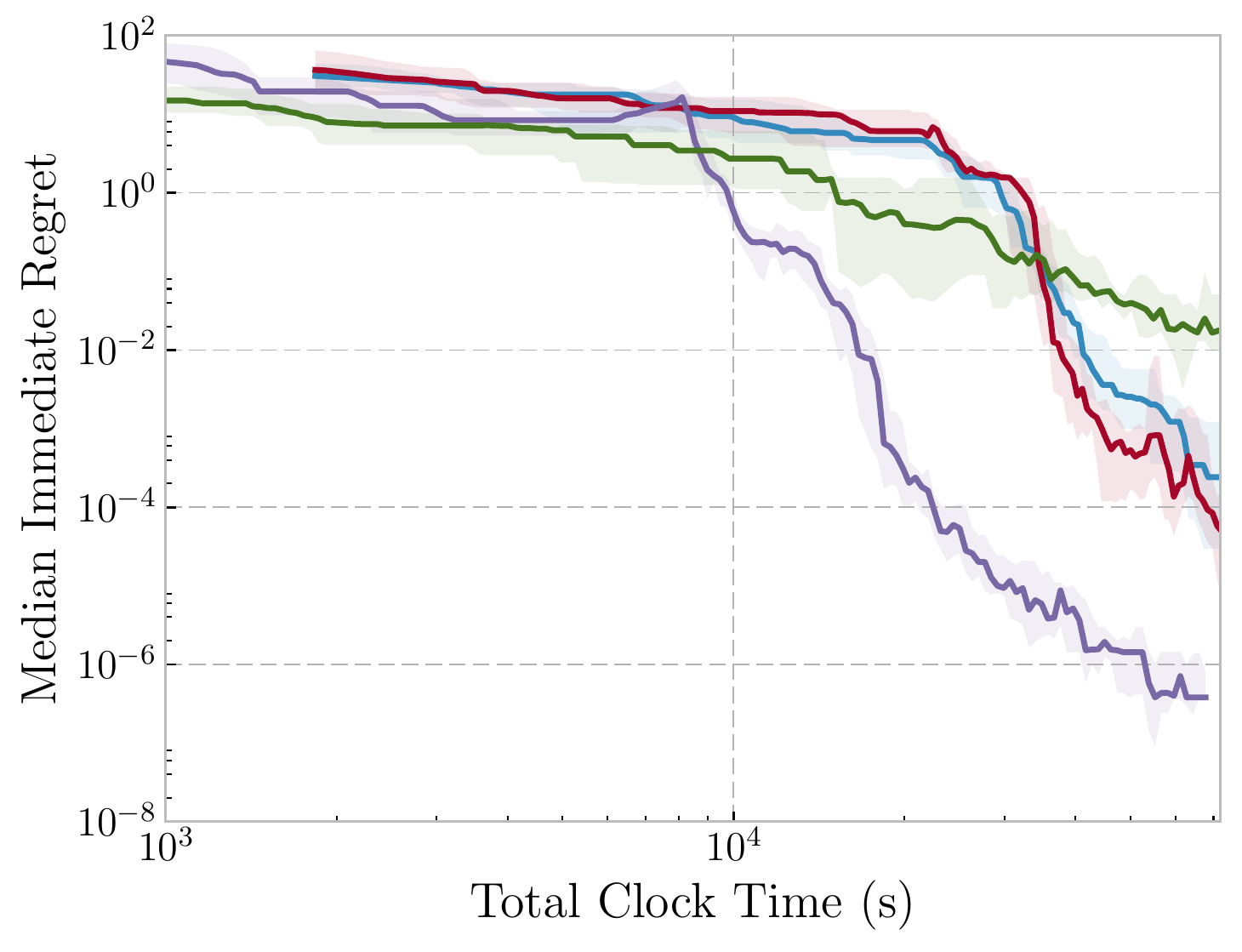}
\includegraphics[width=0.32\textwidth]{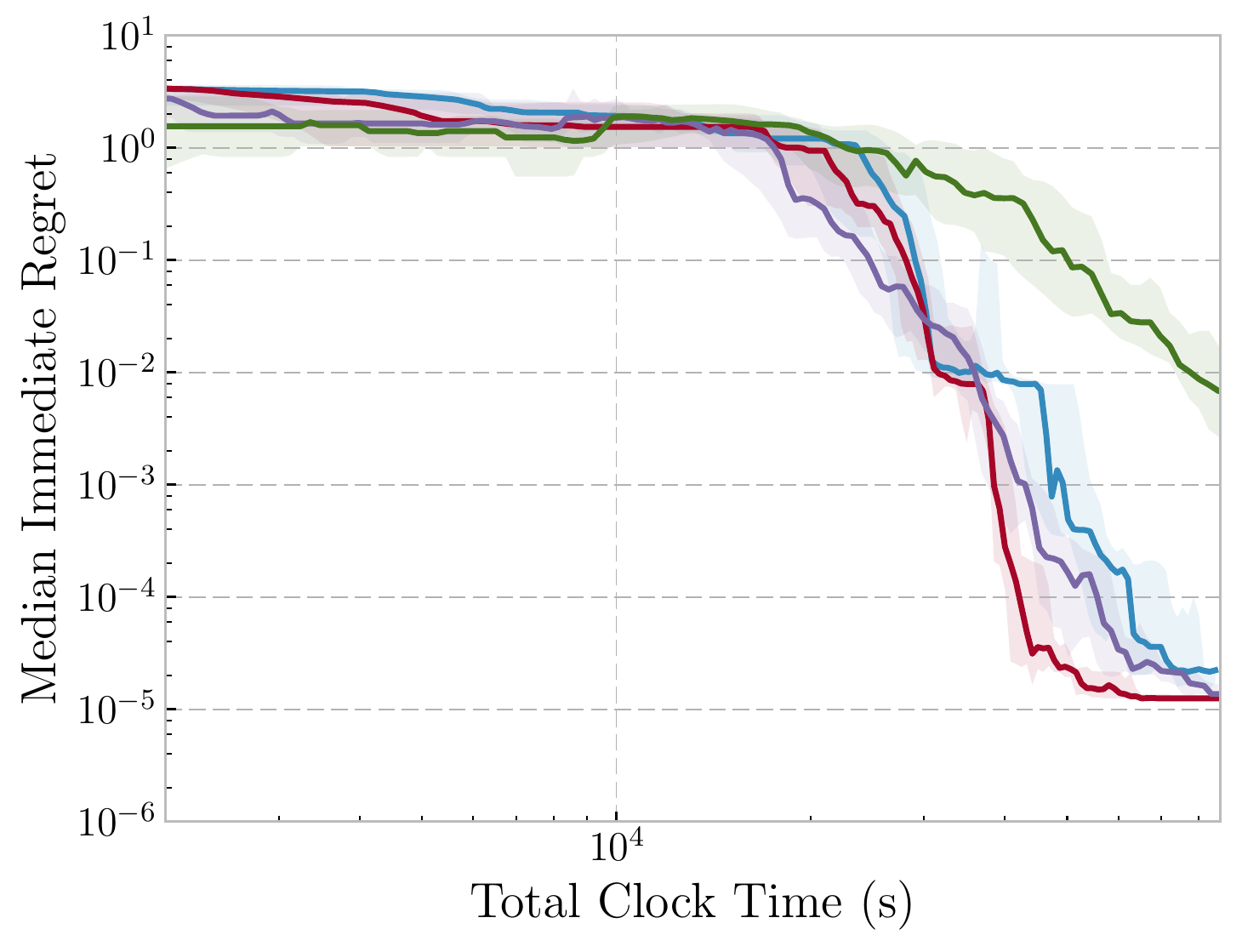}
\includegraphics[width=0.32\textwidth]{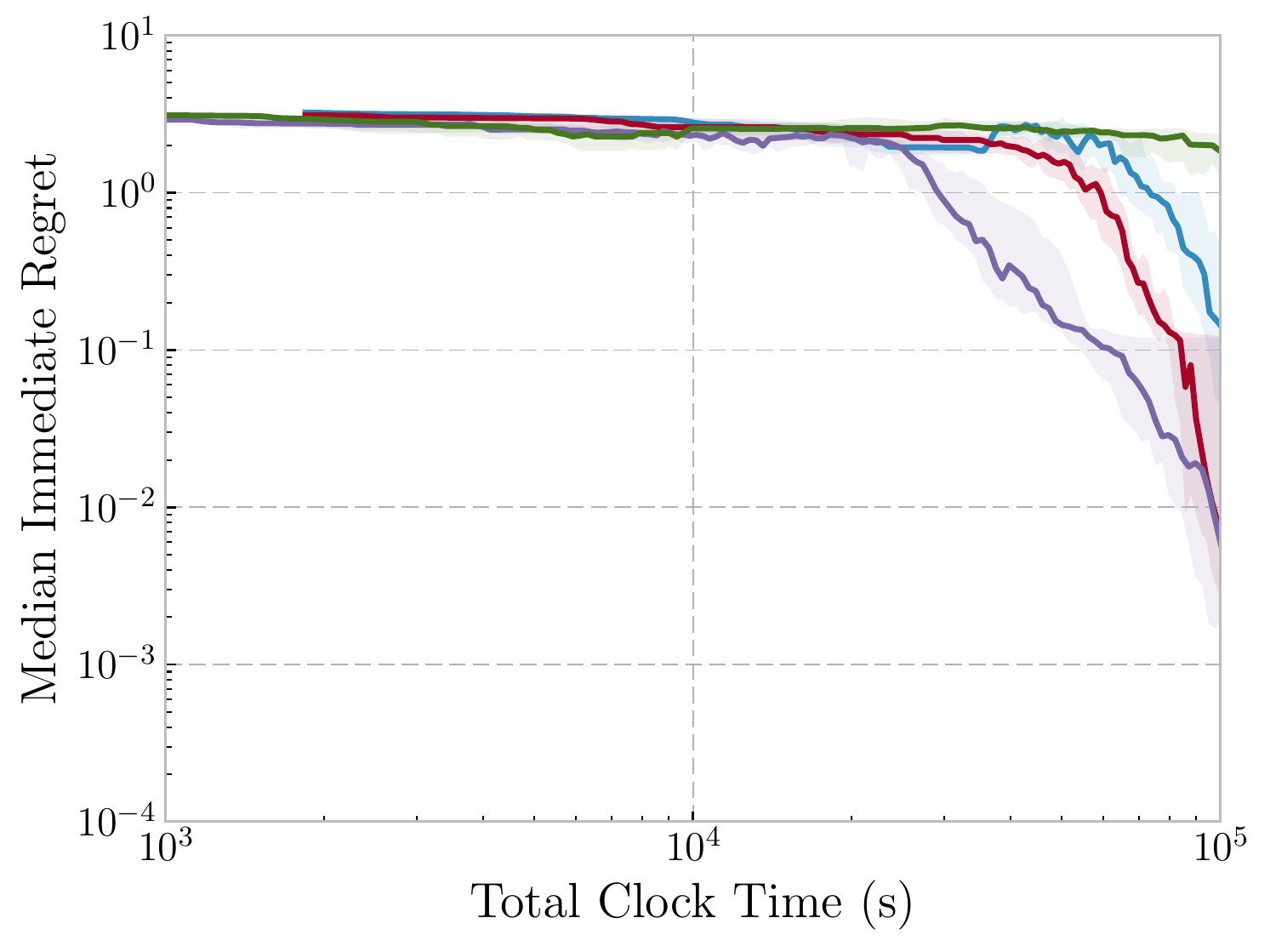}
\caption{Performance of EnvPES (purple), PES (red), Expected Improvement (blue) and FABOLAS (green) with modifications as noted on the Branin (left), Hartman 3D (middle) and Hartman 6D (right) test functions with less expensive evaluations available under a linear shift from the true objective. The performance is shown against evaluation time for the objective (top) and including overhead due to the acquisition function (bottom). The median and interquartile range (shaded) of 20 runs are shown. Note that while FABOLAS has good performance with evaluation cost the high overheads detract from this substantially when included in the cost.}
\label{testfns}
\end{figure*}

We test our method on a selection of common objectives. The results are shown in Figure \ref{testfns}. Following the approach of \cite{swersky2013multi} we use a linear shift of the objective for the lower cost evaluations: in our case the shift is continuous rather than discrete. The cost imposed is of quadratic form rising from two minutes, as the cheapest available, up to thirty minutes for the full objective.
Performance of EnvPES matches or exceeds the other methods shown on each test. Considering only the cost of evaluation, EnvPES and FABOLAS have similar performance, but, acknowledging the real time to optimize the acquisition function, the lower overheads of EnvPES allow better performance.

\subsubsection{MNIST classifer hyperparameters}
\begin{figure}
\centering
\includegraphics[width=\columnwidth]{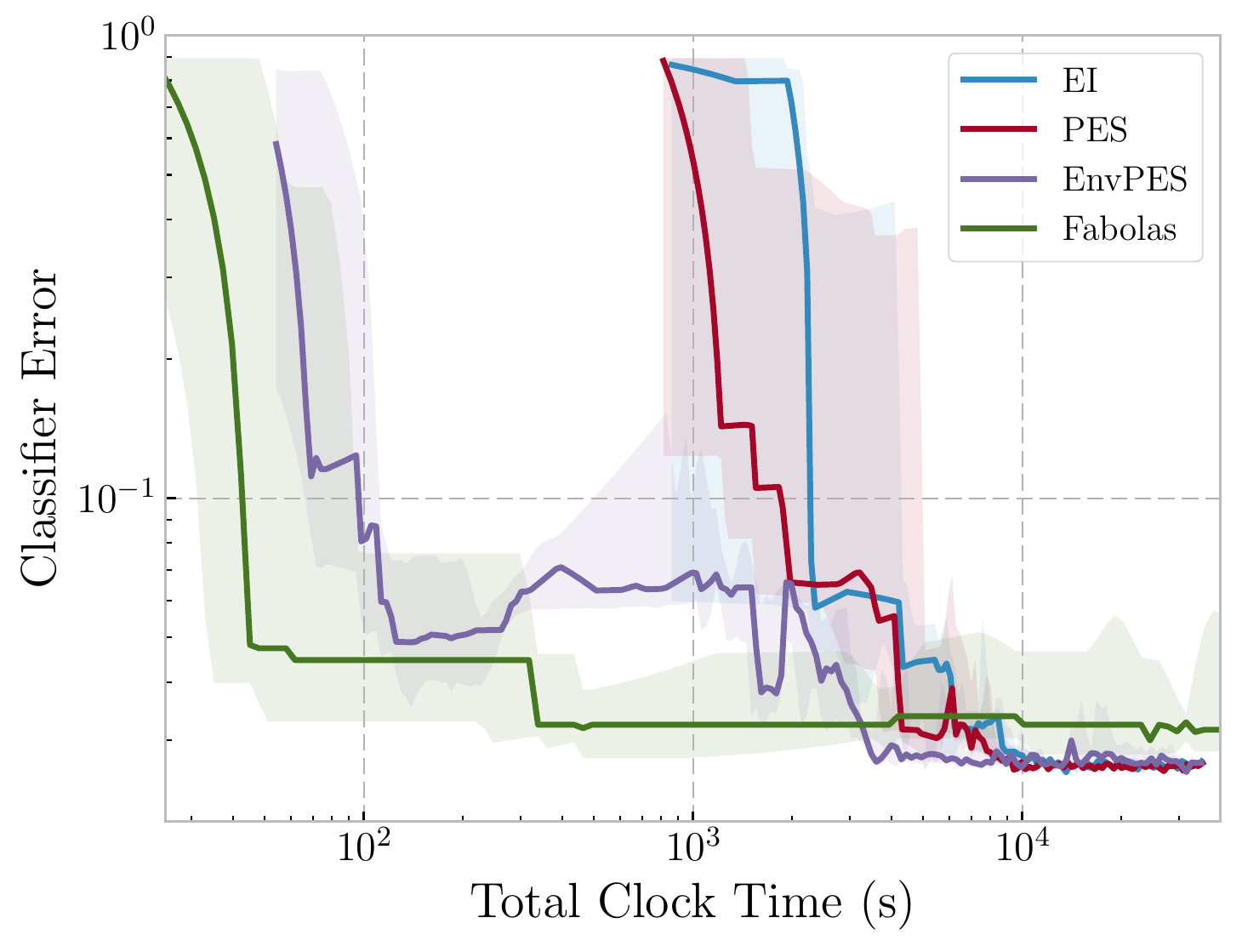}
\caption{Performance of EnvPES (purple), PES (red), Expected Improvement (blue) and FABOLAS (green) finding the best hyperparameters for a support vector machine classifying the MNIST dataset. The median and interquartile range (shaded) of seven runs are shown. Here we have used the original form of FABOLAS.}
\label{mnist}
\end{figure}

Finding the best parameters for a classifier is a common problem in machine learning. We optimize the error penalty and kernel lengthscale hyperparameters of a SVM classifier on the the MNIST dataset. We allow the dataset size to vary from $100$ to $10000$, which incurs a cost of around five minutes on the full dataset using a standard laptop. As shown in Figure \ref{mnist}, we are able to achieve superior performance to the existing methods. Both our method and FABOLAS are able to achieve low values faster than methods not making use of the environmental variable. However, due to the high overhead cost, FABOLAS is then slow to make further improvement.

\subsubsection{GP Kernel Parameter Fitting}
\begin{figure}
\centering

\includegraphics[width= \columnwidth]{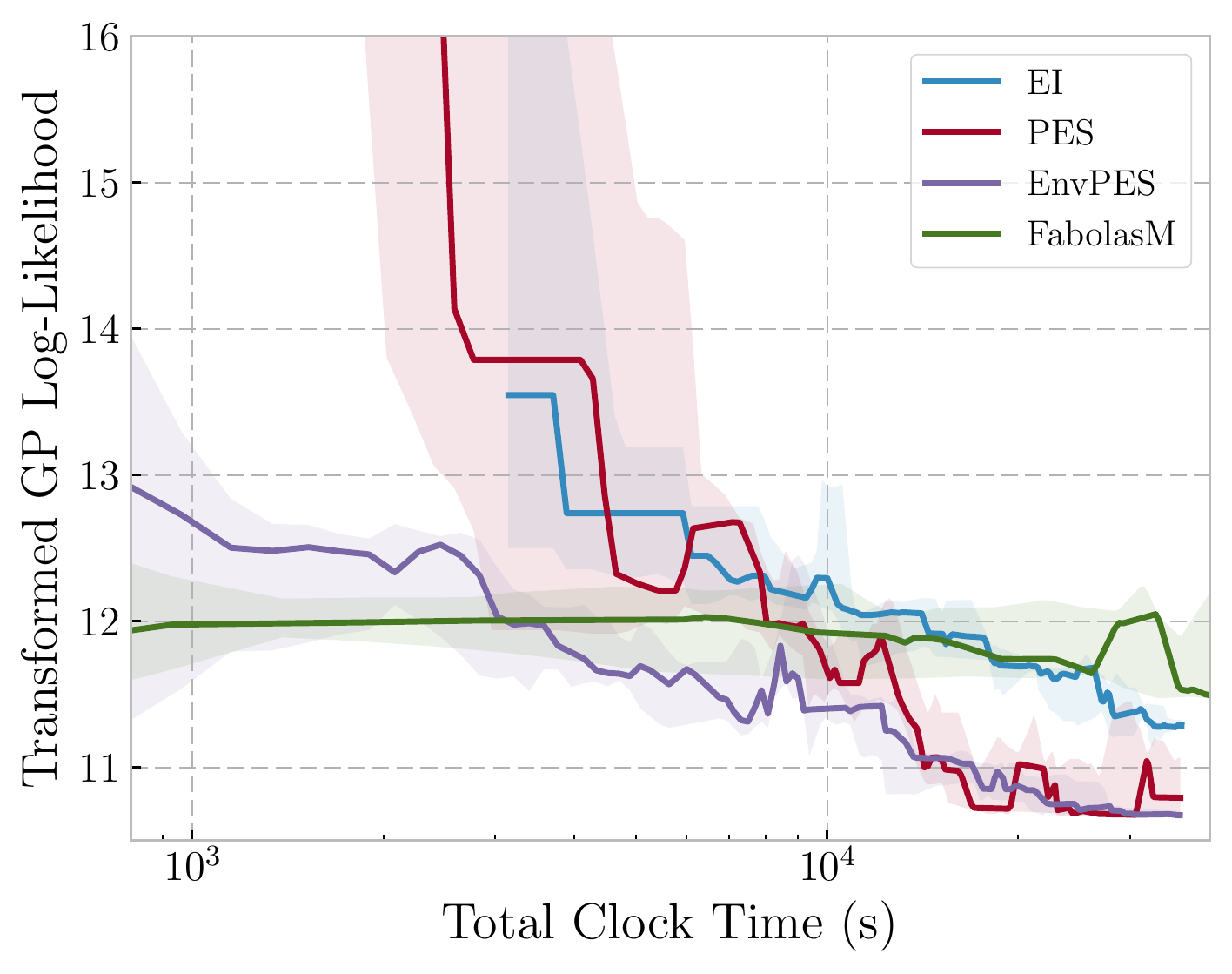}

\caption{Performance of EnvPES (purple), PES (red), Expected Improvement (blue) and FABOLAS (green) minimizing the negative log-likelihood of kernel hyperparameters for a Gaussian Process on UK power data. The median and interquartile range (shaded) of ten runs are shown.}
\label{powerfit}
\end{figure}

Fitting the hyperparameters of a Gaussian Process, our final common machine learning problem, has evaluation cost that scales cubically with the number of datapoints. We train a GP with a Mat\'ern 5/2 kernel on freely available half hourly time series data for UK electricity demand for 2015\footnote{www2.nationalgrid.com/UK/Industry-information/Electricity-transmission-operational-data/Data-explorer}. Evaluation of this objective with the full dataset again typically incurs a cost of around ten minutes. EnvPES is able to evaluate the log-likelihood of random subsets down to $n_{\text{sub}}=0.02N$ of the full dataset. We adjust the log-likelihood by $g(y,n_{\text{sub}}) = -y\frac{n_{\text{sub}}}{N} \,\text{if} \,y<-1,\, 1+\log(-y\frac{n_{\text{sub}}}{N}) \,\text{else}$, which is monotonic, smooth and continuous with respect to the true log-likelihood but reduces the absolute value of large negative likelihoods and normalizes subsets to the value at the full dataset. EI and PES are only able to use the full dataset. As shown in Figure \ref{powerfit} we are again able to achieve to achieve low values much earlier than methods not making use of the environmental variable, and are faster in real-time convergence than FABOLAS due to reduced overheads.

\section{Conclusion}

We have proposed a novel acquisition function based on Predictive Entropy Search for use in variable cost Bayesian Optimization. We further introduce a novel sampling strategy, applicable to both ES and PES, which makes our implementation more computationally efficient by providing a closer approximation to $p(x_*)$ while having substantially lower cost than slice sampling. We have also proposed an alternative method for evaluating the performance of Bayesian Optimization methods. Bringing these together we demonstrate a practical Bayesian Optimization algorithm for variable cost methods and have show that we are able to match or exceed the performance of existing methods on a selection of synthetic and real world applications.
\FloatBarrier

\bibliographystyle{icml2018}

\end{document}